\pdfoutput=1

\documentclass[11pt]{article}

\usepackage{acl}

\usepackage{times}
\usepackage{latexsym}
\usepackage{tabularx}
\usepackage{arydshln}

\usepackage[T1]{fontenc}

\usepackage[utf8]{inputenc}

\usepackage{microtype}
\usepackage{graphicx}
\usepackage{amsmath}
\usepackage{amssymb}
\usepackage{natbib}
\usepackage{amsthm}
\usepackage{booktabs}
\usepackage{multirow}
\usepackage{algorithm}
\usepackage{algorithmic}
 \usepackage{bm}
 \usepackage{multirow}
 \usepackage{color}

\title{DISK: Domain-constrained Instance Sketch for Math \\ Word Problem Generation}

\author {
        Tianyang Cao \textsuperscript{\rm 1,2},
        \ Shuang Zeng \textsuperscript{\rm 1,2},
        \ Xiaodan Xu \textsuperscript{\rm 1,2},
        \ Mairgup Mansur  \textsuperscript{\rm 3} and
        Baobao Chang \textsuperscript{\rm 1,2}\footnotemark[1]   \\
    \textsuperscript{\rm 1}  Key Laboratory of Computational Linguistics, Peking University, MOE, China \\
    \textsuperscript{\rm 2}  School of Software and Microelectronics, Peking University, China \\
    \textsuperscript{\rm 3}  Sogou Technology Inc.            \\
     \{ctymy,zengs,diane1968\}@pku.edu.cn,\ maerhufu@sogou-inc.com,\ chbb@pku.edu.cn
}

\begin{document}
\maketitle

\begin{abstract}
    A math word problem (MWP) is a coherent narrative which reflects the underlying logic of math equations. Successful MWP generation can automate the writing of mathematics questions. Previous methods mainly generate MWP text based on inflexible pre-defined templates. In this paper, we propose a neural model for generating MWP text from math equations. Firstly, we incorporate a matching model conditioned on the domain knowledge to retrieve a MWP instance which is most consistent with the ground-truth, where the domain is a latent variable extracted with a domain summarizer. Secondly, by constructing a Quantity Cell Graph (QCG) from the retrieved MWP instance and reasoning over it, we improve the model's comprehension of real-world scenarios and derive a domain-constrained instance sketch to guide the generation. Besides, the QCG also interacts with the equation encoder to enhance the alignment between math tokens (e.g.,  quantities and variables) and MWP text. Experiments and empirical analysis on educational MWP set show that our model achieves impressive performance in both automatic evaluation metrics and human evaluation metrics.

\end{abstract}
\section{Introduction}
\begin{figure}
 \centering
 \includegraphics[width=1.0\linewidth]{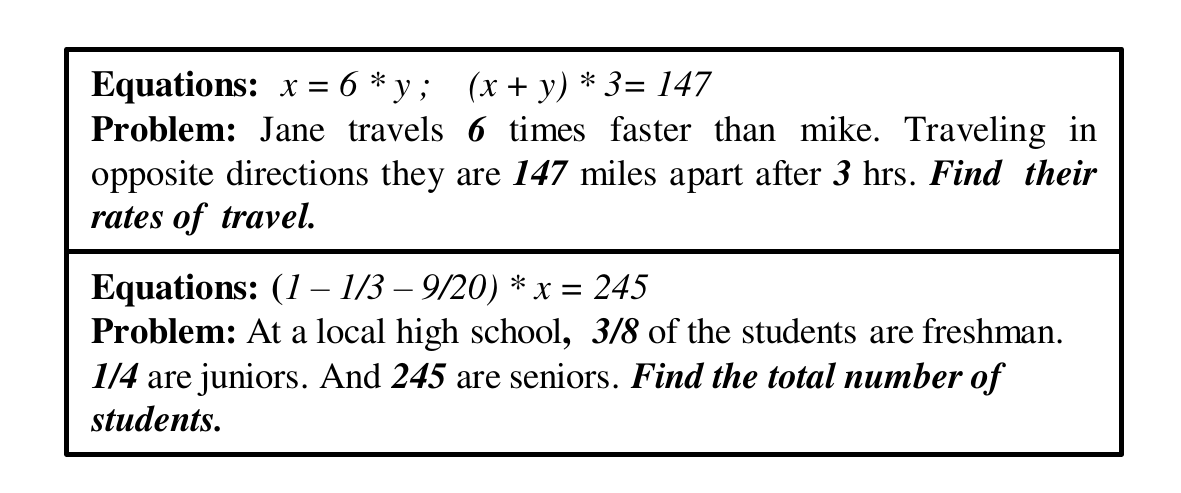}
 \caption{Two examples selected from the MWP generation dataset.}
 \label{table:runing-examples}
\end{figure}

Text generation has been broadly studied as an important task in the field of natural language processing. It aims to generate natural language text that is fluent, readable and faithful to the original input. Recent text generation studies mainly focus on the data-to-text generation, which produces textual output from structured data such as tables of records or knowledge graphs (KGs) \citep{puduppully2019data-to-text,chen-etal-2019-enhancing,gong-etal-2019-enhanced,zhao-etal-2020-bridging}. 
In this paper, we focus on a relatively new type of data-to-text generation task: generating Math Word Problems (MWP) from equations \citep{zhou2019towards}, which does not seem to have been fully studied by the community. Figure~\ref{table:runing-examples} shows two examples of this task. We aim to automatically generate coherent narratives which reflect the computational relationship within given equations. Successful math word problems generation has the potential to automate the writing of mathematics questions given equations to be solved. It can alleviate the burden of school teachers and further help improve the teaching efficiency.

\newcommand{\tabincell}[2]{\begin{tabular}{@{}#1@{}}#2\end{tabular}}

\begin{figure}[htbp]
\centerline{\includegraphics[width=1.0\linewidth]{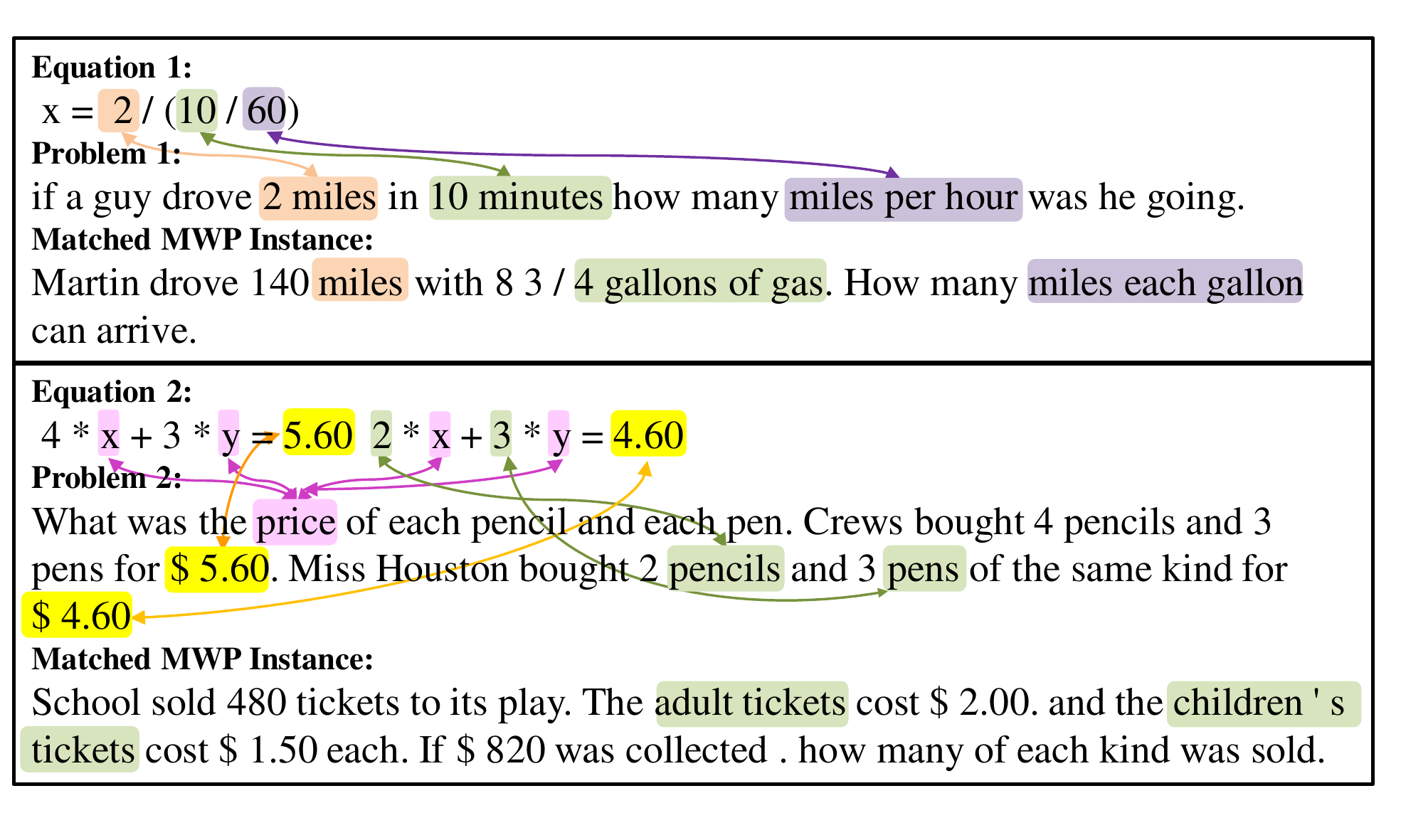}}
\caption{Two examples for illustration of the model. The corresponding MWP instance of each example is selected by a domain-dominated retriever with the participation of domain knowledge. The soft alignments between the math tokens and the corresponding MWP text spans are labeled in the same color. 
}
\label{fig1}
\end{figure}

However, different from other data-to-text generation tasks, 
generating MWP text from abstract math equations is much more challenging. 
Firstly, 
an equation can be expressed by different MWP texts which differ in topic, style or grammar, known as one-to-many pattern. Take the Equation 1 in Figure~\ref{table:runing-examples} for example, ``$x=6*y$'' can be expressed as ``Jane travels 6 times faster than Mike.'', but it is also okay to express it as ``The price of oranges is six times the apples.''. So when grounding the input abstract math equations into a specific math problem, it is hard for a model to decide which scenes to choose for generation. Secondly, the math tokens in equations and natural language text in problems are from completely different symbolic representation space. So this gap increases the difficulty of establishing alignments between math tokens and natural language words, as shown in Figure~\ref{fig1}. Such issue also confuses the generator thus makes generation process uncontrollable. 

\par To overcome these challenges, we propose a novel MWP generator with \textbf{D}omain-constrained \textbf{I}nstance \textbf{SK}etch (\textbf{DISK}). We point out that MWP generation task can benefit from instance sketch restricted by 
domains. For two examples in Figure~\ref{fig1}, DISK first utilizes a domain-dominated retriever to select an instance from a set of candidate MWP text. This instance can be regarded as to constrain the scene with which the math problem text to be generated would be related. Then, DISK produces a refined instance sketch from this instance to specify certain patterns for generating MWP text, using a Quantity Cell Graph (QCG) constructed from the instance. This graph reflects the backbone of the instance with entities and actions related to quantities as nodes. We conduct reasoning over QCG using Graph Convolutional Network \citep{DBLP:conf/iclr/KipfW17} to extract the final instance sketch, where math tokens can be contextualized with corresponding attributes and predicates via interaction between QCG nodes and equations. Finally, the model can generate MWP text with the refined instance sketch using a sequence generator.

Our contributions can be summarized as follows:
\begin{itemize}
    \item We propose a domain-constrained instance sketch guided MWP text generation model, in which the domain information correspoding to the MWP text is automatically induced.
    \item Our model generates the instance sketch via Quantity Cell Graph enhanced encoding, it also contextualizes the math tokens with corresponding attributes and predicates via interaction between QCG nodes and equations.
\end{itemize} 

Experiments show our model can generate more fluent and domain consistent MWP text, with promising performance improvement over strong baselines.

\section{Related Work}
\par \textbf{MWP Generation:} Early MWP generation methods are mostly template-based, including Answer Set Programming (ASP) \citep{polozov2015personalized}, schema and frame semantics \citep{Singley,Deane}. With the development of deep learning framework, \citet{zhou2019towards,wang-etal-2021-math} generate problem text given equation templates and keywords, where the keywords are extracted from the golden MWP via heuristic rules. Their model is learned with Seq2seq in an end-to-end manner and integrates features of templates and keywords in the decoding phase. Their model, however, requires keywords from the golden answer as input when testing, which is unavailable in real scenarios. Another work \citet{DBLP:journals/corr/abs-2010-06196} adopts the external commonsense based knowledge graph (CSKG) to generate topic relevant sentences, while this method only considers the cases of linear equations and needs annotated topics for each equation. So this method is time-consuming and limited. \citet{DBLP:conf/icann/CaoZZMC21} incorporates topic controlling and commonsense enforcement in MWP generation. 

\noindent \textbf{Data-to-text Generation:} Data-to-text generation transforms structured data into descriptive texts \citep{DBLP:journals/nle/Siddharthan01,DBLP:journals/jair/GattK18}. Recent works have brought great promising performance to several data-to-text generation tasks, e.g., \citet{puduppully2019data-to-text,puduppully2019data-to-text1,gong2019table-to-text,wiseman2017challenges} focus on report generation; \citet{chisholm2017learning,lebret2016neural} target at biography generation; \citet{zhao-etal-2020-bridging,DBLP:conf/ijcai/GaoWHX20} generate texts from a set of RDF triples considering structural information. Previous works have also designed content selection and text planning models to determine what to say and how to say \citep{puduppully2019data-to-text,perez-beltrachini-lapata-2018-bootstrapping}.

\noindent \textbf{Retrieval-based Generation:} The methods similar to our instance-based generation are the skeleton-then-response frameworks which are popular in dialogue response generation \citep{cai-etal-2019-skeleton,DBLP:conf/aaai/0006WHWL019,yu-jiang-2021-expanding,DBLP:conf/ijcai/CaiCSZY20}. These models usually treat the input text as a query and the similar query along with its response in databases is then retrieved with Information Retrieval (IR) systems. However, they rely on difference between the input query and retrieved query to identify informative words in the retrieved response, which can not be employed in our task since it is meaningless to measure the similarity between equations.

\section{Model}

\begin{figure*}[h]
\centerline{\includegraphics[width=0.9\textwidth]{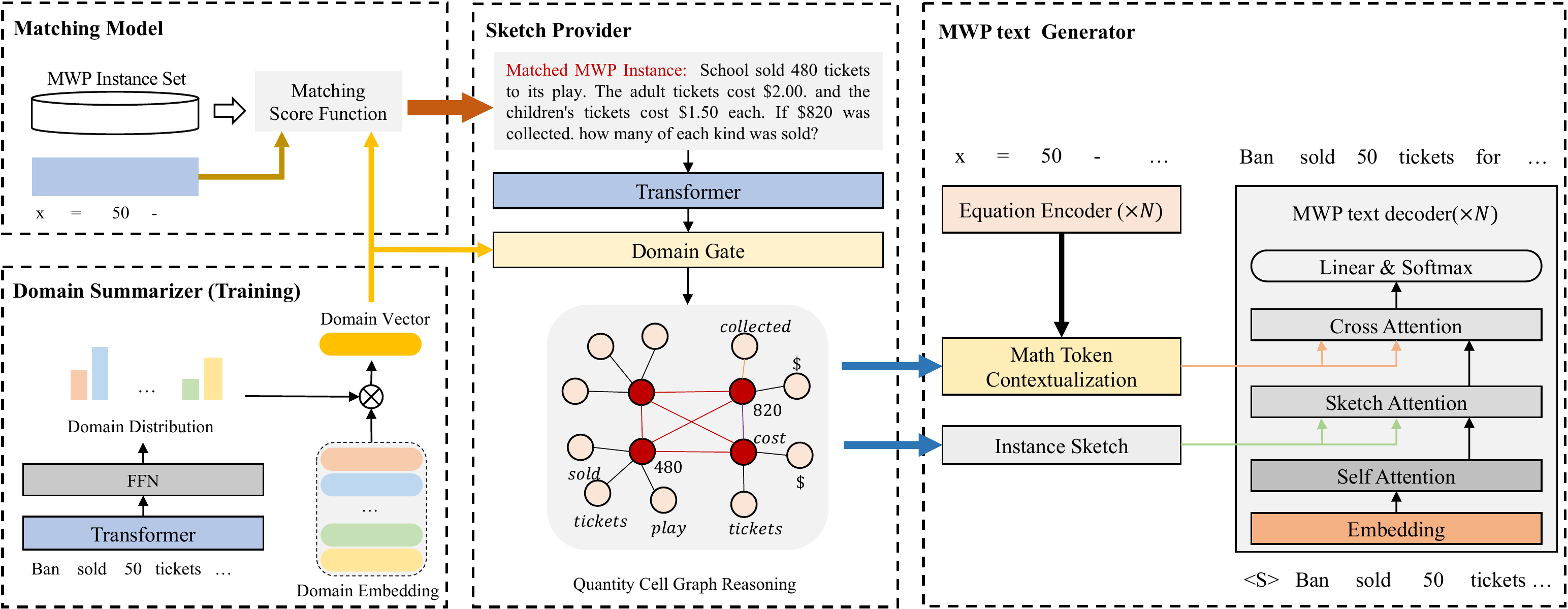}}
\caption{The diagram of proposed DISK. The text-domain vector is the summation of domain embeddings. First, the matching model predicts the most consistent MWP instance based on the equation tokens together with the text-domain vector. Then, the sketch provider learns to refine the retrieved MWP instance with a domain gate and the Quantity Cell Graph, it also contextualizes the equation representation to help the model understand the alignment between equations and MWP text. Finally, the generator consumes both the instance sketch and contextualized equation representation for generating.
}
\label{fig2}
\end{figure*}

The overview of our DISK is depicted in
Figure~\ref{fig2}. Our model follows a three-stage procedure: Firstly, given the input equations and a text-domain vector which is extracted by the \textbf{Domain Summarizer}, the \textbf{Matching Model} retrieves a most similar MWP instance in database by jointly measuring equation-text matching score and domain-text matching score. Secondly, the \textbf{Sketch Provider} enriches the original representation of the instance to yield a refined instance sketch, it filters out excessive information considering domain constraint and helps the model to understand quantity relationship by applying Graph Neural Network (GNN) over the Quantity Cell Graph (QCG). Thirdly, \textbf{Text Generator} generates MWP text via utilizing both the math equation contextualized by QCG and the refined instance sketch based on an encoder-decoder architecture.

\subsection{Domain Summarizer}
The domain summarizer takes the MWP text $y=\left\{y_i\right\}_{i=1}^L$ with length $L$ as input and its goal is to collect underlying domain information in the MWP text, which contributes to instance retrieving. 
To this end, inspired by \citep{inproceedings,unknow}, we assume $K$ latent domains are depicted by the MWP text with different importance $\beta_i$ and the text-domain vector can be expressed as the weighted sum of $K$ trainable domain vectors.

\par We start by encoding the MWP text into a sequence of vectors via a transformer block:
\begin{equation}\label{eq1}
    \left[\bm{h}_1;\bm{h}_2;...,\bm{h}_L\right] = Encoder_P(\left[\bm{y}_1;\bm{y}_2;...;\bm{y}_L\right])
\end{equation}
\noindent where $\left[;\right]$ denotes concatenation operation. We denotes $\left[\bm{h}_1;\bm{h}_2;...;\bm{h}_L\right]$ as $\bm{H}\in \mathbb{R}^{L\times d}$, $d$ is the embedding size. A global attention is applied to the output of the transformer:
\begin{equation}\label{eq2}
   \bm{h}_a = \sum_{i=1}^L \alpha_i \bm{h}_i
\end{equation}

where the attention weight $\alpha_i=\text{softmax}(\bm{h}_i\bm{W}^a \overline{\bm{h}})$, $\overline{\bm{h}}  = \frac{1}{L}\sum_{i=1}^L \bm{h}_i$.  A nonlinear transformation is utilized to fuse the encoded MWP text into $K$ domain variables:
\begin{equation}
    \tilde{\bm{D}} = tanh(\bm{W}^D_1( \bm{H}\bm{W}^D_2+\bm{b}^D_2)+\bm{b}^D_1)
\end{equation}
$\tilde{\bm{D}}\in \mathbb{R}^{K\times d}$ and parameters $\bm{W}^D_1\in \mathbb{R}^{K\times L}$, $\bm{W}^D_2\in \mathbb{R}^{d\times d}$. Each row vector in $\tilde{\bm{D}}$ corresponds to a different domain contained in MWP text. Such process can be treated as a soft clustering and we hope each domain expresses its unique aspect. Similar to \citet{article}, we employ an auxiliary loss function to restrict the derived $K$ domain representation to be orthogonal with each other:
\begin{equation}
    \mathcal{L}_D = ||\tilde{\bm{D}}\tilde{\bm{D}}^T - \bm{I}_{K\times K}||
\end{equation}
$\bm{I}_{K\times K}$ is an identity matrix. We then map $\tilde{\bm{D}}$ and $\bm{h}_a$ to a domain distribution with an attention mechanism \cite{bahdanau2014neural}:
\begin{equation}
    \beta_i = \text{Softmax}(\bm{v}^T tanh(\bm{H}^d\bm{h}_a+\bm{U}^d\tilde{\bm{D}}_{i,:}))
\end{equation}
where $\bm{v},\bm{H}^d,\bm{U}^d$ are learnable parameters. 
$\beta_i$ indicates the domain distribution of the given MWP. Our model then learns a trainable domain embedding $\bm{E}\in \mathbb{R}^{K\times d}$ and use $\left\{\beta_i\right\}_{i=1}^K$ to compute the text-domain vector over $\bm{E}$: $\bm{h}_d = \sum_{i=1}^K \beta_i \bm{E}_{i,:}$. Note that the domain summarizer only works during training process. During test, we enumerate each discrete domain vector in $\bm{E}$ to be fed into the matching model, which will be illustrated later.

\subsection{Matching Model}
The matching model aims to match one MWP instance from the training corpus which is most consistent with the given equation. Intuitively, incorporating the domain variable helps our model better recognize MWP text with similar domain grounding to the golden problem text, since it’s difficult to infer from the equation only. Thus it's rational to combine the text-domain vector and the math equation to retrieve an additional instance.
\par Our matching model ranks all MWP texts from a pre-defined set $P=\left\{P_1,P_2,...,P_{|P|}\right\}$, and returns the most consistent one with given equation-MWP pair $(x,y)$, where $P$ is prepared by uniformly sampling from the training corpus. The text-domain vector $\bm{h}_d$ and equation embedding $\left\{\bm{x}_i\right\}_{i=1}^N$ serve as input, where $\bm{x}_i$ is the sum of corresponding token embedding and type embedding, here type embedding is incorporated to distinguish quantities, numbers and operations in math equations. Each text $P_i=p^i_1p^i_2...p^i_{|P_i|}$ is encoded into context representation $\left\{\bm{u}^i_j\right\}_{j=1}^{|P_i|}$ through transformer blocks (denoted as $Encoder_Q$). However, the dataset provides no supervision for the matching model, we then annotate the golden labels by ranking the BERTScore \citep{bert-score} between each candidate MWP text and the ground-truth MWP, i.e., $\text{BERTScore}(P_i,y) \  1\leq i \leq |P|$, getting the top one $P_{l^*}$ as the selected instance. \\
\textbf{Equation to MWP Matching}: Equation to MWP matching score is measured in token-level. 
Firstly, we encode $\left\{\bm{x}_i\right\}_{i=1}^N$ into $\bm{C}=\left\{\bm{c}_i\right\}_{i=1}^N$ via a transformer block, $\left[\bm{c}_1,\bm{c}_2,...,\bm{c}_N\right] = Encoder_E(\left[\bm{x}_1,\bm{x}_2,...,\bm{x}_N\right])$. A nonlinear function is then used to compute the correlation score between the $j$-th token in the equation and the $k$-th token of $P_i$:
\begin{equation}
    \gamma_{i,j,k}= g_1(\bm{c}_j)\cdot g_2(\bm{u}^i_k)
\end{equation}

where $g_1(\cdot),g_2(\cdot)$ are both multi-layer perceptrons (MLP). 
Next, we aggregate token level relevance to determine text-level score:
\begin{equation}\label{eq7}
     s_{em}(i) = \mathop{mean}\limits_{j,k}\gamma_{i,j,k}
\end{equation}

\textbf{Domain to MWP Matching}: We interact the text-domain vector $\bm{h}_d$ with the context representation of $P_i$: $\left\{\bm{u}^i_j\right\}_{j=1}^{|P_i|}$ to obtain the domain to MWP matching score. Similar to \eqref{eq2}, we apply a global attention to calculate the summation of $P_i$, which we denote as $\bm{h}_p^i$. We then compute the domain vector to the $i$th MWP text relevance vector via a bilinear transformation:

\begin{equation}\label{eq8}
    \bm{r}_i = \bm{h}_d\bm{W}^r\bm{h}_p^i
\end{equation}
$\bm{W}^r \in \mathbb{R}^{d\times d' \times d}$ is a parameter. Finally, we combine \eqref{eq7} and \eqref{eq8} to produce the normalized distribution over $|P|$ candidate MWP texts:

\begin{equation}\label{eq9}
    s(i) = \text{Softmax}(s_{em}(i)+\bm{w}^r\bm{r}_i) \ i\in\left[1,|P|\right]
\end{equation}
where $\bm{w}^r \in \mathbb{R}^{d'}$ is a learnable parameter. With the label $l^*$ annotated, we add a cross entropy loss to supervise the result of the matching model:
\begin{equation}
    \mathcal{L}_M = - \text{log}(s(l^*))
\end{equation}
For inference, the top one MWP text with the highest matching score is selected to be fed into the next sketch provider.

\subsection{Sketch Provider}
The sketch provider aims to generate the instance sketch by making soft modification to the MWP instance representation, since the generative model should capture underlying patterns contained in the instance, not simply copy the instance. We achieve this goal in two aspects: 1) we add a domain gate to refine tokens that has high relevance with the domain information, 2) we incorporate the Quantity Cell Graph to enable our model to better understand complex question scenarios while maintaining those spans semantically aligned with equation tokens.
\par Firstly, for the encoded representation of $P_{l^*} :\left\{\bm{u}_i\right\}_{i=1}^{|P_{l^*}|}$ (which has been processed by $Encoder_Q$ in the matching model), we employ a soft gate controlled by the domain vector $\bm{h}_d$ to better flow the important context in the original matched text:
\begin{align}\label{eq11}
    \bm{q}_i &= \sigma(\bm{W}^q\left[\bm{h}_d;\bm{u}_i\right])  \nonumber \\
    \bm{u}_i' &= \text{tanh}(\bm{W}^Q\bm{u}_i)\odot \bm{q}_i \ \ 1\leq i\leq |P_{l^*}|
\end{align}
\textbf{Quantity Cell Graph Constructing and Reasoning} 
Targeting at better exploiting the retrieved MWP instance, we should enrich the instance encoding with quantity relationship information, as well as effectively guide the alignment between abstract equation tokens and MWP text tokens. Inspired by \citet{zhang-etal-2020-graph}, we introduce the Quantity Cell Graph (QCG), whose nodes contain a subset of tokens in the the MWP text related to numerical values. As is shown in Fig~\ref{fig2}, a Quantity Cell Graph is composed of a set of Quantity Cells (QC): $QCG = \left\{QC_1,QC_2,...,QC_m\right\}$, where $m$ denotes the number of quantities in the matched MWP instance $P_{l^*}$. Each cell $QC_i$ can be expressed as $\left\{v^q_i\right\} \bigcup \left\{v^a_{i,1},v^a_{i,2},...\right\}$, where $v^q_i$ is the $i$th quantity token and $v^a_{i,1},v^a_{i.2},...$ is the corresponding attributes or predicates. We resort to Dependency Tree\footnote{https://demo.allennlp.org/dependency-parsing} and Constituency Tree \footnote{https://demo.allennlp.org/constituency-parsing} to extract attributes related to each quantity token. Details can be found in Appendix ~\ref{graph}.
We argue that the extracted tokens are salient properties related to quantities and show explicit alignment with the input equations. With the nodes in the Quantity Cell Graph mentioned above, we add an edge between two nodes if 1) they are both quantity nodes, 2) one is the quantity node and another is the attribute node belonging to it. Next, we initialize the node representation of the QCG by concatenating the corresponding output of $Encoder_Q$ and its POS tag embedding, which is denoted as  $\bm{S}^0=\left\{\bm{s}_k\right\}_{k=1}^{|G|}$, $|G|$ is the node number of the QCG. Graph Convolutional Network \citep{DBLP:conf/iclr/KipfW17} is applied to capture the dependencies between QCG nodes:
\begin{equation}
\bm{S}^{l+1} = ReLU(GCN(\bm{S}^l,\bm{A})) 
\end{equation}
where $\bm{S}^l$ is the node representation after the $l$-th layer, $\bm{A}\in \left\{0,1\right\}^{|G|\times|G|}$ is the adjacency matrix. \\
\textbf{Graph2Text Augmentation} After graph network reasoning, we need a fusion block to propagate the aggregated information of the QCG back to the text representation $\bm{U}'=\left\{\bm{u}'_i\right\}_{i=1}^{|P_{l^*}|}$. To locate the position of each node in the original matched text, we establish a binary matrix $\bm{M}\in \left\{0,1\right\}^{|P_{l^*}|\times |G|}$, where $\bm{M}_{ij}=1$ if the $i$-th token in the MWP instance is the $j$-th node in the graph. As each column of $\bm{M}$ corresponds to one quantity node or attribute node in the QCG, we update $\bm{u}'_i$ with a GRU module if the $i$-th token participates in the QCG reasoning:
\begin{equation}
    \tilde{\bm{U}} = GRU(\left[\bm{U}';\bm{M}\bm{S}^{\mathcal{L}}\bm{W}^{U}\right])
\end{equation}
where $\bm{S}^{\mathcal{L}}$ is the output of the last GCN layer and $\bm{W}^{U}$ is a parameter matrix. $\tilde{\bm{U}}$ is treated as the output instance sketch.\\
\textbf{Math Token Contextualization} As mentioned before, the attribute words related to quantities are beneficial to help the model identify the soft alignment pattern between the math equation tokens and retrieved MWP instance. The encoded vector sequence of the input equations $\bm{C}$ attends to the QCG nodes to receive graph information:
\begin{align}\label{eq15}
    \bm{G} &= \text{Softmax}(\bm{C}\bm{W}^G\bm{S}^{\mathcal{L}})  \nonumber \\
    \overline{\bm{C}} &= ReLU(\bm{G}\bm{S}^{\mathcal{L}})
\end{align}
We then calculate two update gate $\bm{f}=\sigma(\bm{W}^g\left[\bm{C};\overline{\bm{C}}\right])$ and $\bm{g}=\sigma(\bm{W}^f\left[\bm{C};\overline{\bm{C}}\right])$, which combines $\bm{C}$ and $\overline{\bm{C}}$ to obtain contextualized equation representation $\tilde{\bm{C}}$:
\begin{align}
    \tilde{\bm{C}} &= \bm{g} \odot \bm{C} + (\bm{1} - \bm{g}) \odot tanh(\bm{W}^Z\left[\bm{C};\bm{f} \odot \overline{\bm{C}}\right]) 
\end{align}

\subsection{MWP Generator}
The MWP generator maps the math equation tokens $x_1x_2...x_N$ to the MWP text, we employ a transformer based encoder-decoder architecture. Here our encoder shares its parameters with $Encoder_E$ in matching model to capture common attention features among them. To enable the decoder to rewrite the domain-constrained instance sketch produced by the sketch provider in a fine-grained manner, we insert an extra sketch attention layer between the original self-attention layer and cross-attention layer. It aggregates details of the sketch by attending to output of sketch provider $\tilde{\bm{U}}$:
\begin{equation}
    \bm{H}' = MultiHeadAttn(\textbf{Q}: \bm{H}_p,\textbf{K}:\tilde{\bm{U}},\textbf{V}:\tilde{\bm{U}})
\end{equation}
where $\bm{H}_p$ is the hidden state coming from the previous layer. Residual connection and layernorm is also added after the sketch attention. For the cross attention layer, the new representation $\tilde{\bm{C}}$ coming from Math Token Contextualization module is used as both the key and value. The hidden state of the last decoder layer is projected to vocabulary distribution and predicts the next token. The domain vector $\bm{h}_d$ is directly fed into the MWP text decoder and serves as the first input embedding (instead of the embedding of a start symbol $<$S$>$). The generation loss can then be modeled as:
\begin{equation}
    \mathcal{L}_G = -\sum_{t=1}^L\text{log}\ p(y_t|y_{<t},\left\{x_i\right\}_{i=1}^N,\tilde{\bm{U}},\bm{h}_d)
\end{equation}
$y_{<t}$ is the tokens generated before the $t$-th step.

\subsection{Model Training}
For training, we combine the three loss terms mentioned above and the total loss becomes:
\begin{equation}
    \mathcal{L}_{total} = \mathcal{L}_D + \mathcal{L}_M + \mathcal{L}_G
 \end{equation}
For inference, our model traverses over all $K$ possible latent domain vectors in $\bm{E}$ and generates $K$ candidate MWP texts $Y^1,Y^2,...,Y^K$, among which the problem text with the maximum log likelihood score is chosen as the final output.

\section{Experiments}
\subsection{Dataset}

\begin{table*}[htbp]
\scriptsize
\centering
\scalebox{0.8}{\begin{tabular}{p{50pt} p{40pt} p{45pt} p{45pt} p{45pt} p{40pt} p{40pt}  p{40pt}}
\toprule

 \textbf{Model}   & \textbf{BLEU} & \textbf{ROUGE-L} & \textbf{BERTScore} & \textbf{METEOR} & \textbf{Dist1}($\%$) & \textbf{Dist2}($\%$) & \textbf{NR}($\%$)  \\
\midrule
Seq2seq    &  2.59   & 20.25   &82.98 &  18.51  &  14.56    & 34.99  & 47.60\\
SeqGAN  & 2.62    & 19.22   &82.56  &17.63   & 12.96     &30.02   & 44.00  \\

DeepGCN &  3.04   & 20.94   &83.07  & 19.48  &16.81      &45.17    & 49.21 \\
Transformer & 3.14    &21.84    &83.81  &20.26   & 12.94     &43.51   &44.84  \\
DualCG & 3.60    & 21.43   &83.99  & 20.63  &  15.47    &  46.01 & 40.97   \\
$\text{BART}_{large}$ & 4.15   & 22.26 & \textbf{86.35}   & 22.30 & 12.77 & 46.76 &  43.47    \\
\midrule
 \tabincell{l}{DISK} & \textbf{5.84}    &  \textbf{28.49}  &85.01 & \textbf{27.16}   &  \textbf{15.56}    &  \textbf{50.41} &  52.62  \\
\tabincell{l}{w/o DG} &  5.33   &  27.36  & 84.90  &25.33   &  10.74    & 35.62  &49.37   \\
  w/o QCG   & 4.86    &  26.87  & 84.96 & 26.56   & 13.88     & 45.19  & 55.15  \\
  w/o MTC   &  5.14   & 28.08   & 84.62  & 26.16   &  13.63   & 43.90  & 55.51  \\
 
 w/o CS & 4.59   & 26.96     &84.27 & 25.32   & 12.53   & 43.67    & \textbf{57.68}  \\
 \bottomrule
\end{tabular}}
\caption{ Automatic evaluation results of different models in MWP generation dataset. NR is the abbreviation for Number Recall. }
\label{auto}
\end{table*}

Our dataset is based on Dolphin18K \citep{huang-etal-2016-well} crawled from Yahoo. Since \citet{huang-etal-2016-well} only releases a subset of Dolphin18K (3154 examples), which is insufficient for a modern data-driven generation model. So we reuse the python scripts provided by \citet{huang-etal-2016-well} to crawl and collect extra data, then the size of dataset is extended to 14943 examples. We conduct some data preprocessing by deleting those equation-problem text pairs whose problem text length is longer than 45 tokens or less than 10 tokens. Finally 9643 samples are preserved. More detailed statistics of the dataset are listed in Appendix ~\ref{data}.

\subsection{Baselines}
We compare DISK against the following models. 1) \textbf{Seq2seq} \citep{bahdanau2014neural}: Seq2seq is first proposed for machine translation task. In this paper, we implement Seq2seq with attention mechanism and copy mechanism. 2) \textbf{SeqGAN} \citep{yu2016seqgan:}: SeqGAN fuses the advantage of reinforcement learning (RL) and Generative Adversarial Network (GAN). It achieves improvements over strong baselines in both text generation and music generation tasks. 3) \textbf{DeepGCN} \citep{Guo}: Math equations can be converted into a pre-ordered expression tree and MWP generation can then be naturally modeled as graph-to-sequence learning. 4) \textbf{Transformer} \citep{vaswani2017attention}: The state of-the-art model in several text generation tasks. 5) \textbf{DualCG} \citep{DBLP:conf/nips/WeiL0FJ19}: In this paper we employ DualCG to integrate equation to MWP generation and MWP to equation solving in a unified framework.  6) We also compare our model with the vanilla \textbf{BART} \citep{lewis-etal-2020-bart}, which is a strong pretrained model using the standard seq2seq Transformer architecture. We fine-tune BART on our MWP dataset. Note we do not use retrieval-based baselines since they mostly require IR system, while it's unsuitable to treat math symbols as the query.

\subsection{Automatic Evaluation}

We compare different methods using BLEU (average of BLEU-1 and BLEU-2) \citep{papineni2002bleu:}, ROUGE-L \citep{lin2004rouge:}, METEOR \citep{DBLP:conf/acl/BanerjeeL05}, BERTScore \citep{bert-score}, which is an advanced evaluation metric for text generation based on contextual embedding, Dist-1, Dist-2, which indicates the proportion of different unigrams (bigrams) in all unigrams (bigrams), Number Recall, which is used to measure how many numbers in problem text are correctly copied. We report the performance of all models in terms of automatic evaluation in Table~\ref{auto}.
\par We also conduct ablation studies and the results are also reported in Table~\ref{auto}. The setting is as follows: 1) w/o DG: We remove the domain gate in the Sketch Provider. 2) w/o QCG: We remove both the QCG reasoning and the Math Tokens Contextualization block. The encoded MWP instance, after being processed by the domain gate, is directly fed into the generator in this case. 3) w/o MTC: The model without Math Token Contextualization. 4) w/o CS: The model without the instance sketch, i.e., the whole Matching Model and Sketch Provider are removed and only the Domain Summarizer and the Generator are preserved.

\par It can be observed that 1) our proposed model significantly outperforms the strongest DualCG in BLEU, ROUGE-L and BERTScore, respectively. It yields higher results in most language quality metrics even when compared with $\text{BART}_{large}$. Besides, our model also improves the informativeness and diversity of generated MWP. 2) simply letting the MWP decoder attend to the retrieved MWP (w/o QCG) will degrade the performance by 16.78\%, 5.69\%, 0.06\% in BLEU, ROUGE-L and BERTScore, respectively, which proves the Quantity Cell Graph can guide the generator to understand the quantity relationship and better exploit the retrieved MWP instance. Moreover, as an intermediate result, we report the capacity of the matching model in Appendix ~\ref{match}.

\subsection{Performance on Different Types of Equations}

Table~\ref{subset} shows the performance on different subsets of the MWP generation dataset (divided by the number of variables). We can see the proposed method outperforms baselines by a large margin in all subsets. Intuitively, the more variables the equation contains, the more imperatively the generation process needs the guidance of instance sketch. It's easy to show our model obtains more absolute gain in More Than Three-VAR subset than One-AVR or Two-VAR ones. 
\begin{table}[htbp]
\small
\centering
\scalebox{0.7}{\begin{tabular}{c l c c c}
\toprule
\multirow{3}{*}{\textbf{One-VAR}} &  & \textbf{BLEU} & \textbf{ROUGE-L} & \textbf{BERTScore} \\
\cline{2-5}
 & \textbf{DualCG}  & 2.88  &  19.99  & 83.59    \\
 \cline{2-5}
 & \textbf{Trans} & 2.18  & 19.43 & 83.45  \\
 \cline{2-5}
 & $\textbf{BART}_{large}$ & 3.16  & 19.83 & 85.94 \\
 \cline{2-5}
 & \textbf{DISK} & 3.87  & 27.17  & 84.45   \\
 \midrule
 \multirow{3}{*}{\textbf{Two-VAR}} &  & \textbf{BLEU} & \textbf{ROUGE-L} & \textbf{BERTScore} \\
 \cline{2-5}
 & \textbf{DualCG}  &3.75  &  23.05  & 84.41    \\
 \cline{2-5}
 & \textbf{Trans} & 3.97  & 22.82 & 84.52  \\
 \cline{2-5}
 & $\textbf{BART}_{large}$ & 4.77  & 25.26 & 87.00 \\
 \cline{2-5}
 & \textbf{DISK} &  6.33 &  29.03 & 85.51   \\
 \midrule
 \multirow{3}{*}{\tabincell{c}{\textbf{More Than} \\ \textbf{Three-VAR}}} &  & \textbf{BLEU} & \textbf{ROUGE-L} & \textbf{BERTScore} \\
 \cline{2-5}
 & \textbf{DualCG}  & 2.00 & 17.74  &84.33     \\
 \cline{2-5}
 & \textbf{Trans} & 3.33  &21.10  & 83.15  \\
 \cline{2-5}
 & $\textbf{BART}_{large}$ & 1.86  & 17.16 & 84.97 \\
 \cline{2-5}
 & \textbf{DISK} & 4.59   & 25.63  & 84.26   \\
 \bottomrule
\end{tabular}}
\label{subset}
\caption{Performance on different subsets on our MWP generating dataset. Trans is short for Transformer.}
\label{subset}
\end{table}

\begin{table}[h]
\setlength{\abovecaptionskip}{-5pt}
\setlength{\belowcaptionskip}{-10pt}
\small
\centering
\scalebox{0.7}{\begin{tabular}{l p{24pt} p{24pt} p{24pt} p{24pt} p{24pt} p{24pt} }
\toprule
 & \multicolumn{2}{c}{\textbf{Fluency}} & \multicolumn{2}{c}{\textbf{Coherence}} &\multirow{2}{*}{\textbf{S1}(\%)} & \multirow{2}{*}{\textbf{S2}(\%)} \\
\cline{2-5}
& score & $\kappa$ & score & $\kappa$ &   &   \\
\midrule
DISK & \textbf{4.00} & 0.413 & \textbf{4.08} &0.497 & \textbf{36} & \textbf{56} \\
\midrule
Seq2seq & 3.78  & 0.256  & 3.48 &0.483  &  23 & 34 \\
SeqGAN & 3.75 & 0.305 & 3.28 &0.520 & 20 & 40 \\
 
DeepGCN & 3.61 & 0.295 & 3.55 & 0.494 & 29 & 52 \\
Transformer & 3.80 &0.333 & 3.53 &0.421 & 20 &45 \\
DualCG  & 3.88  &0.346  & 3.66 & 0.455  & 28   &  53   \\
$BART_{large}$ & 3.56  & 0.398 & 3.73   & 0.454 & 31  &  52 \\
\bottomrule \\
\end{tabular}}
\caption{Human evaluation results: comparison between the proposed model and baseline models. }
\label{table:human-evaluation}
\end{table}

\subsection{Human Evaluation}
To better measure the actual generation quality, we recruit three human annotators to judge the quality of different models, which includes four aspects listed as follows. 1) \textbf{Fluency}: Fluency mainly judges whether the problem text is fluent, i.e., whether some grammar errors occur in generated MWP. 2) \textbf{Coherence}: Coherence weights if the problem text is consistent in text-level. 3) \textbf{Solvability-1 (S1)}: As our target is a math word problem, we should pay attention to whether the problem text can be solved, i.e., in what percentage we can set up the same (or equivalent) equations and solve them according to the generated problem text. 4) \textbf{Solvability-2 (S2)}: Solvability-2 is a more relaxed criterion compared with Solvability-1, it only requires the text produced is a valid math problem and could be solved regardless what equations could be set.

\begin{table*}[!htbp]
\scriptsize
	\centering
	\scalebox{0.75}{\begin{tabularx}{20.1cm}{X}
		
		\toprule
		
		 \textbf{Equ}: $equ : 250 + 400 = x \ equ : 1625/x = y$ \newline \textbf{MT}: \textcolor{purple}{2 vehicles traveling} different directions. same \textcolor{purple}{start point} and time. one vehicle is \textcolor{purple}{60 mph}. the other is \textcolor{purple}{55 mph}. In how many hours will they be \textcolor{purple}{500 miles apart}.
        \newline \textbf{Ours}: \textcolor{blue}{Two cars leave Denver traveling} in opposite directions. One has a \textcolor{blue}{speed of 250 mph} and the other airplane \textcolor{blue}{averages 400 mph}. How many hours will the trip be \textcolor{blue}{1625 miles apart}. (\emph{BLEU}: 15.47) \newline
     \textbf{Seq2seq}: A $<$UNK$>$ of deposit costs \$ 400. 000 a t the end of the year. the total interest is \$ 1625 . 00. What is the total cost of the total.\newline
    \textbf{SeqGAN}: quotient of a certain number is 400. If the number of students in the first 250 is 400. What is the number.\newline
    \textbf{DeepGCN}: The car ran a t an average speed of 400 km per hour faster than the other. If the speed of a 400 mi / h faster. What was the speed of the plane in miles per hour.  \newline \textbf{Trans}: planes went to school a t a speed for the trip takes 250 mph for 400 hours. How long will the plane travel in the trip.  \newline\textbf{DualCG}: Joe received 250 miles for 250 miles . and gas a trip of 250 miles per hour for \$ 400 to the week. He drove 400 miles per hour faster . What was the average speed for the trip. 
		\\
		\midrule
		\midrule
		\textbf{Equ}: $ equ : x + y = 35 \  equ : x / y = 2 / 5$  \newline \textbf{MT}: total of 1600 people work for \textcolor{purple}{a company}. \textcolor{purple}{The ratio of male to female employees} is 3 : 5. How many more \textcolor{purple}{females than males} are there in the company. \newline
		\textbf{Ours}: The \textcolor{blue}{ratio of boys to girls} in at \textcolor{blue}{a certain school} is 5 : 2. If there are total 35 \textcolor{blue}{boys and girls}. how many of each are there. (\emph{BLEU}: 9.09)\newline
        \textbf{Seq2seq}: The school art club is having a exhibit. The ratio of the school paintings are in two parts is 2 / 5 of the number. What is the number ?\newline \textbf{SeqGAN}: A carpet is 3 times as many more than the other. The total value is 3.\newline
        \textbf{DeepGCN}: The ratio of the larger of the two numbers is 35. The ratio of the smaller number of goals and the other is 5 / 2. What are the two numbers.  \newline \textbf{Trans}: Pat . 35 students and 5 questions. If the total of the students are seniors and 2 take both the total. how many of each. \newline
        \textbf{DualCG}: The sum of two numbers is 35. The larger number is 2 less than the smaller number. Find the larger number. \\
		\bottomrule
	\end{tabularx}}
	\caption{Two examples of math word problems generated by different models. Transformer is abbreviate to Trans. Equ and MT represent the equation and the matched MWP instance, respectively. Quantity-related attributes and predicates in the instance that are picked up and rewritten in the generated MWP are colored for better readability. }
	\label{case}
\end{table*}

\par We randomly select 100 generated MWP texts and score them in five grades. We then project the scores to 1$\sim$5, where higher scores indicate better performance. Moreover, we assess the inter-annotator agreement by Cohen's kappa $\kappa$, which reflects the agreement between scores given by different raters. The averaged results are reported in Table~\ref{table:human-evaluation}. We can clearly see that the proposed model performs much better than other models, not only in fluency and coherence, but also in the solvability of generated math word problem.

\subsection{Effectiveness of the Domain Gate and Quantity Cell Graph}
\begin{figure}[h]
 \centering
 \includegraphics[width=1.05\linewidth]{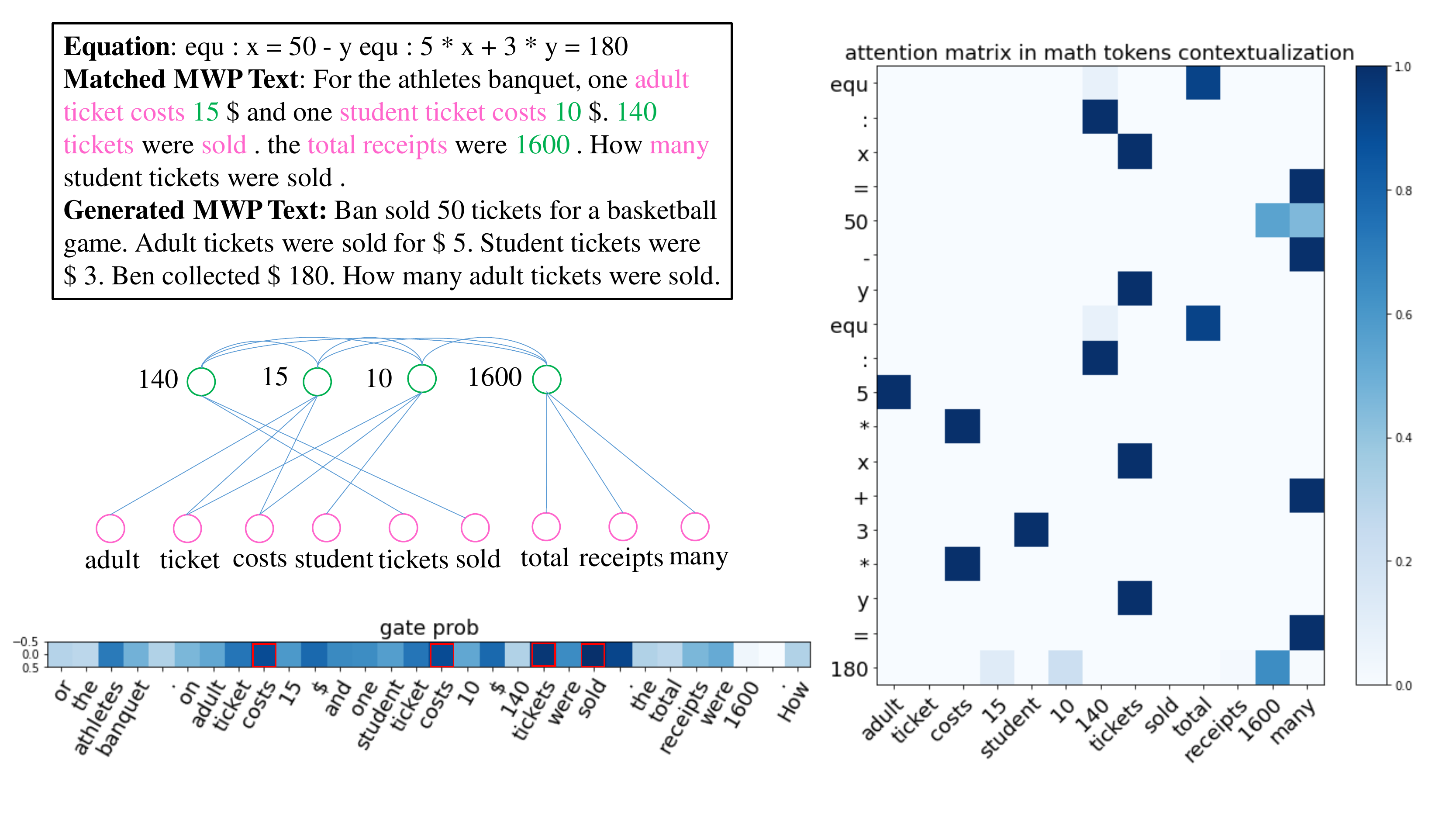}
\caption{Visualization of a test case, which shows 1) the retrieved MWP text and the generated MWP text 2) the extracted QCG from the retrieved MWP text 3) the value of the domain gate on different tokens in \eqref{eq11} 4) the attention matrix between the input equation representation and the QCG, namely, $\bm{G}$ in \eqref{eq15}}
 \label{visual} 
\end{figure}
We show the effectiveness of the domain-dominated soft gate and the Quantity Cell Graph Reasoning through qualitative analysis. Fig~\ref{visual} presents a test case processed by our model. The heatmap in the left lower corner indicates the relevance of each token to the text-domain vector in the matched MWP text. The top-4 tokens are marked with the red box. Words that highlight the characteristic of one certain domain, e.g., ``tickets'', ``sold'', ``cost''... are assigned with higher weight to be fused into the next block. The heatmap on the right hand side presents the probability that each equation token attends to the nodes in the QCG (after normalization). It is easy to show: number ``5'' is aligned with ``adult'', since ``5'' is the price of adult tickets; number ``3'' is aligned with ``student'', since ``3'' is the price of student tickets; both ``$x$'' and ``$y$'' are aligned with ``tickets'', since ``$x$'' and ``$y$'' both imply the number of tickets; ``180'' is aligned with ``1600'' as 1600 is the total receipts in the matched MWP instance and ``180'' also refers to the total sales in our generated text. It's reasonable to believe that the math token contextualization enhances the semantic alignment between math equations and the matched MWP instance.

\subsection{Case Study}

Table~\ref{case} shows two examples in the test dataset generated by different models. Additional examples can be found in Appendix ~\ref{more_case}. We can observe that: 1) DISK gives consistent context in text-level while keeping readability, which verifies it's effective to assign a domain vector to each MWP text. 2) The generated MWP text expresses plausible attributes related to quantities by making an analogy with the matched instance. E.g., in case 2, the matched text discusses the ratio of male employees and female employees, while the MWP generated by our model says ``the ratio of boys to girls''. Besides, it's interesting that though in case 2, the output given by our system receives low BLEU score, it's still a logically reasonable and feasible MWP. So BLEU score may not be suitable for evaluating the performance of MWP generation. According to the above analysis, it is obvious that instance sketch provider improves the informativeness of the given equation via correctly understanding and exploiting the connections among QCG nodes.

\section{Conclusion}
We propose DISK, which introduces latent discrete domains for matching appropriate MWP instance and refines its representation. We also extract Quantity Cell Graph to enhance the sketch-guided generator and help our model better understand math equations in real scenarios. Experimental results on the extended Dolphin18K Dataset show the superiority of our model.

\bibliography{anthology,custom}
\bibliographystyle{acl_natbib}

\clearpage
\appendix













\section{Details on Constructing Quantity Cell Graph}\label{graph}
In this section, we describe the rules for extracting quantity-related attributes as follows.
\begin{itemize}
    \item We consider those tokens which are labeled as \emph{Nouns} or \emph{Verbs} and are within two hops starting from the quantity token in the dependency tree.
    \item We firstly traverse the nodes in the constituency tree starting from the root node in a depth-first manner, and backtracks when the visited node contains no more than $F$ ($F$ is a hyperparameter) leaf nodes. Such operation yields several subtrees and each token in the original text belongs and only belongs to one subtree. We detect the tokens belonging to the same subtree as the quantity token and are labeled as \emph{Nouns} or \emph{Verbs}.
\end{itemize}

\section{Dataset}\label{data}
\textbf{Dataset Information}: Table ~\ref{tab2} provides our data statistics. \\
\textbf{Motivation of Extending Dolphin18K}: MWP solving datasets currently used include Alg514 \citep{kushman-etal-2014-learning}, Dolphin1878 \citep{shi-etal-2015-automatically}, DRAW-1K \citep{upadhyay-chang-2017-annotating}, Dolphin18K \citep{huang-etal-2016-well}. Table \ref{dataset} gives the statistic of these datasets. 
Alg514, Dolphin1878, DRAW-1K are all public available, while neural generation models for generative tasks are usually data-hungry thus equation-MWP pairs in those datasets are insufficient. Though Dolphin18K is a large-scale dataset, only a part of it (3154) are released. Moreover, existing datasets only include a certain type of MWP text, e.g., MWP text for linear equations, which restricts their practical application. We then reuse the python script provided by \citet{huang-etal-2016-well} and acquire 14943 equation-MWP text pairs in total from Yahoo !. Generally, the public available datasets can be treated as the subset of our dataset. Next, we conduct data preprocess as follows, which is beneficial to train the generation model:
\begin{itemize}
    \item We normalize the equations by replacing all the equation variables in each sample to $x,y,z$,... in order, e.g., $u+v+r=100,u-r=10$ is replaced to $x+y+z=100, \ x-z=10$.
    \item We manually correct the wrong spelling words in MWP text.
\end{itemize}
\begin{table}[htbp]
\small
\centering
\scalebox{0.8}{\begin{tabular}{c c c c}
\hline
    & \textbf{Train} & \textbf{Dev} & \textbf{Test} \\
\hline
\textbf{Size}& 7714 & 964 & 965  \\
\hline
\textbf{Equation Length (average)}& 16.69 & 16.23 & 16.63 \\
\hline
\textbf{Problem Length (average)} & 28.90 & 29.64 & 28.74\\
\hline
\textbf{Tokens} &  7445  & 3065 & 2875     \\
\hline
\end{tabular}}
\caption{Statistic of datasets }
\label{tab2}
\end{table}

\begin{table}[h]
\small
\centering
\scalebox{0.75}{
\begin{tabular}{c c c p{25pt} p{25pt}}
\toprule
\textbf{Dataset} & \textbf{Size}  & \textbf{Problem Type} & \textbf{Avg EL} & \textbf{Avg Ops} \\
\midrule
Alg514 & 514 & algebra, linear & 9.67 & 5.69\\
Dolphin1878 & 1878 & number word problems & 8.18 & 4.97\\
DRAW-1K & 1000 & algebra, linear, one-variable & 9.99 & 5.85\\
Dolphin18K &  $18460^{*}$  & algebra, linear,  multi-variable  & 9.19 & 4.96  \\
\midrule
Our Dataset & 14943 & \tabincell{c}{algebra, linear/nonlinear, \\multi-variable} & 16.64  & 6.41
\\
\bottomrule
\end{tabular}}
\caption{Statistics of several existing MWP solving datasets. Avg EL, Avg Ops refer to average equation length and average numbers of operators in equations, respectively. $*$ indicates only 3154 equation-MWP pairs of Dolphin18K are available.}
\label{dataset}
\end{table}

\section{Experimental Settings}\label{setting}
The batch size for training is 32. The vocabulary size is set as 13k. The hidden size for both our model and baseline models is 256. We use 2 layers transformer block in our model and the baseline Transformer model. All multi-head attention layers are implemented with 8 heads. The embeddings are randomly initialized and are trained together with our model. The domain number is set as $K=25$, however, the results for different values of $K$ are also presented in this paper. The size of the candidate MWP set prepared for retrieving is $|P|=500$. For extracting the QCG with constituency parser, the hyper-parameter is set as $F=5$ and the graph network is stacked for 2 layers. To calculate the BERTScore, we use the tool released by the author on Github \footnote{https://github.com/Tiiiger/bert\_score}. We train all models for 40 epoch. To prevent overfitting, we set the dropout probability as 0.2. We use the Adam optimizer \citep{kingma2014adam:} with the learning rate $lr=0.0005$ and momentum $\beta_1=0.9$, $\beta_2=0.999$.

\section{Impact of Different Domain Numbers}
\begin{figure}
 \centering
 \includegraphics[width=0.8\linewidth]{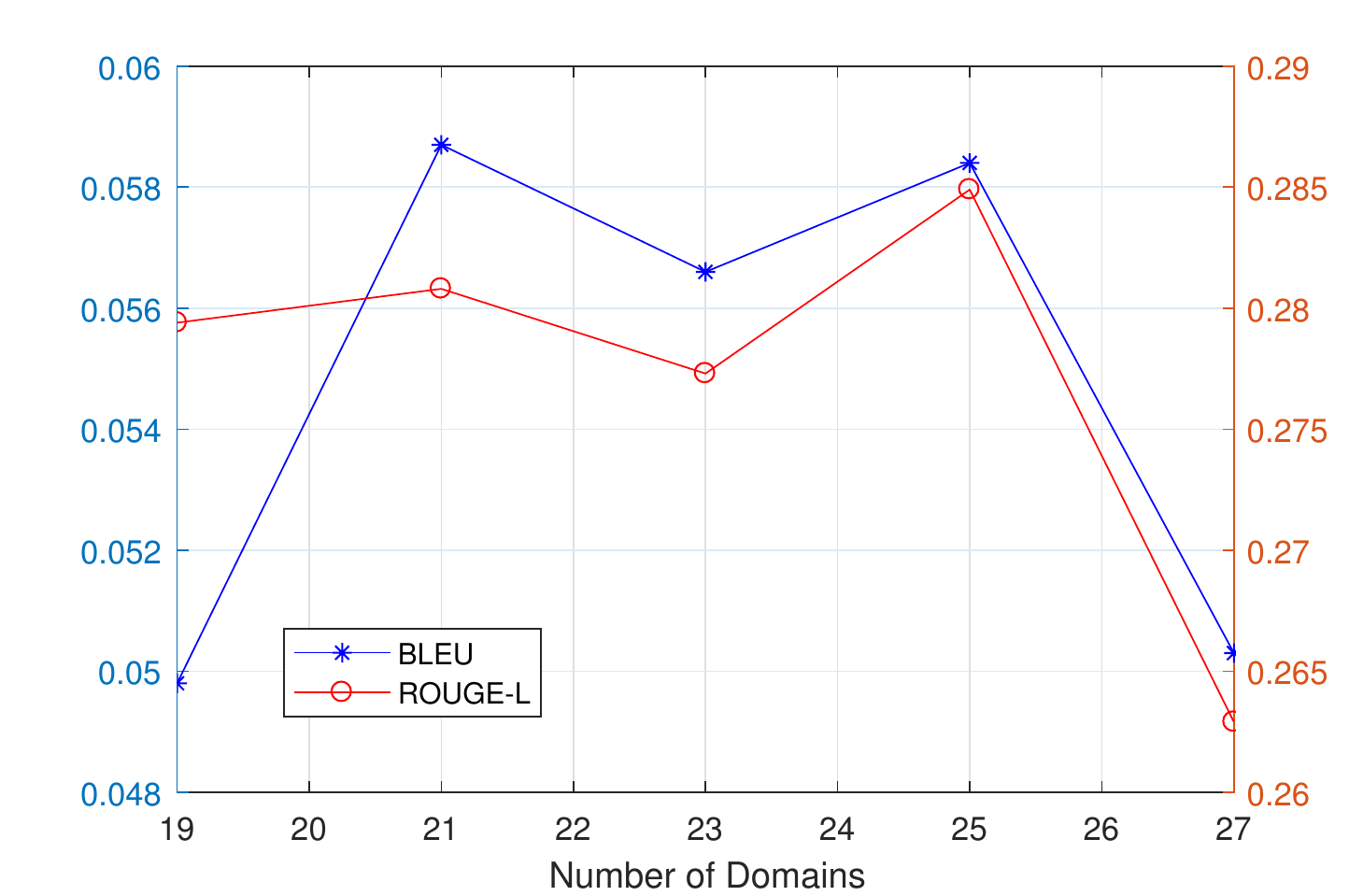}
 \caption{Performance with different domain numbers on the test dataset.}
 \label{domain}
\end{figure}

Fig~\ref{domain} compares the fluctuation of BLEU and ROUGE-L when the number of domains changes from 19 to 27. The proposed model receives consistent improvement compared against the baselines with different numbers of domains, while the peak value appears when $K=21$ or $K=25$. Even though $K$ is set to 19 or 27, our model still exceeds baselines, which demonstrates its generalization capacity.
\par
\begin{figure}
 \centering
 \includegraphics[width=0.8\linewidth]{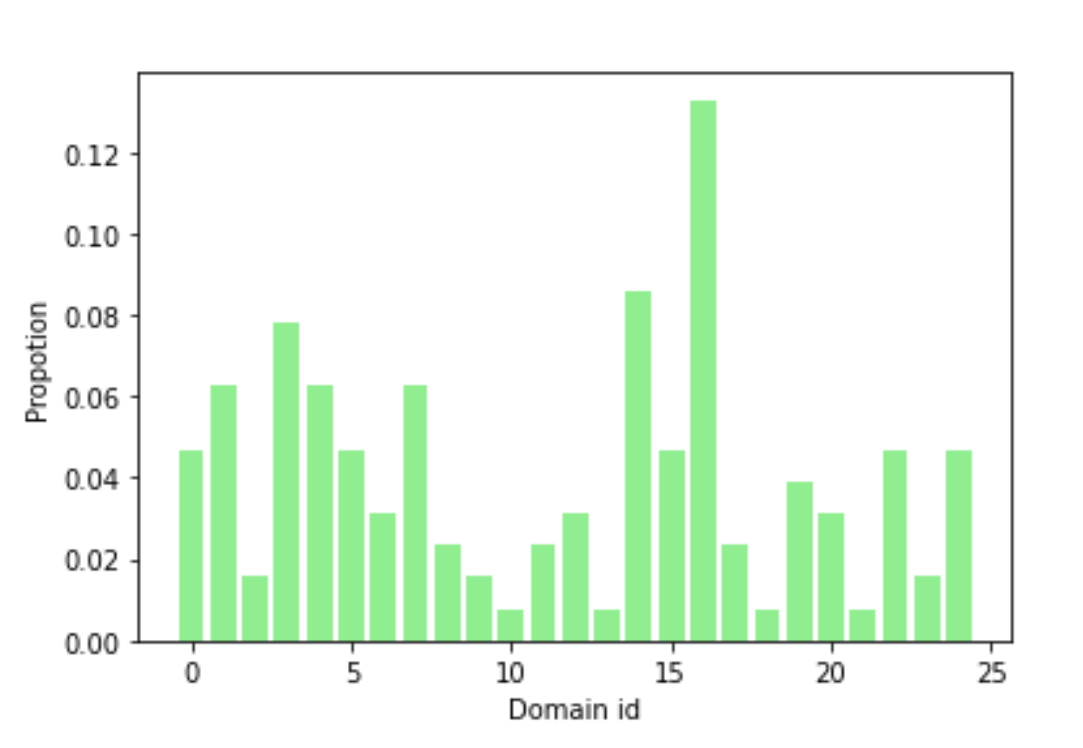}
 \caption{Proportion of test cases which is conditioned on each domain.}
 \label{percentage}
\end{figure}

 Moreover, one interesting problem is whether each domain plays a role during test. To this end, we investigate the percentage of output MWP text which is conditioned on each domain in the whole test set, the result is reported in Fig~\ref{percentage}. We can find our model doesn't lead to ``domain collapse'', i.e., all cases are generated from the same domain, since the distribution of domains are generally balanced.

\section{Effectiveness of the Domain Gate and Quantity Cell Graph}\label{match}
We also conduct ablation study to measure our model's ability of understanding the retrieved MWP instance and properly exploiting the quantity relationship implied in it. As is shown in Table~\ref{instance-pred}, we report the semantic matches in deep between the retrieved MWP instance and the generated MWP. When we discard the Quantity Cell Graph module or the Math Token Contextualization module, the relevance between the MWP instance and generated MWP both drop, which indicates the interaction between quantity-related attributes and enriching equation tokens with concrete scenarios are crucial for performance improvement.
\begin{table}[htbp]
\small
\centering
\scalebox{0.8}{\begin{tabular}{c c c}
\toprule
    & \textbf{METEOR} & \textbf{BERTScore}  \\
\midrule
\textbf{Full Model}&24.09  &85.01   \\
\midrule
\textbf{w/o QCG}& 22.70 & 84.79  \\
\textbf{w/o MTC} &22.86  & 84.28 \\
\bottomrule
\end{tabular}}
\caption{Semantic similarity between the retrieved MWP instance and generated MWP from different models.}
\label{instance-pred}
\end{table}

\section{Example Outputs}\label{more_case}
We show more cases produced by different models in Table~\ref{case} to demonstrate the performance of our proposed model.

\begin{table*}[!htbp]
\small
	\centering
	\scalebox{0.8}{\begin{tabularx}{20.1cm}{X}
		
		\toprule
		
		 \textbf{Equ}: $equ : x + y = 360 \ equ : 0.18 * x + 0.09 * y = 360 * 0.17$ \newline  \textbf{MT}: one can drink is \textcolor{purple}{15\% orange juice}. another is \textcolor{purple}{10\% orange juice}. How many \textcolor{purple}{liters} should be \textcolor{purple}{mixed} in order to get 5 l that is 11\% orange juice. How many liters of the 15\% orange juice should be in the \textcolor{purple}{mixture}. \newline
		\textbf{Ours}: \textcolor{blue}{soybean meal} is \textcolor{blue}{18\% protein}. \textcolor{blue}{cornmeal} is \textcolor{blue}{9\% protein}. How many \textcolor{blue}{pounds} of each should be \textcolor{blue}{mixed} together in order to get \textcolor{blue}{360 lb mixture} that is 17\% protein. (wh)\newline
        \textbf{Seq2seq}: kets to a junior high school play cost \$ 3.50. and student tickets are \$ 360. How many adult tickets were sold.\newline
        \textbf{SeqGAN}: The sum of two numbers is 360 . The sum of the numbers is 360 . What is the number.\newline
        \textbf{DeepGCN}: $<$UNK$>$ has 360 pounds . cornmeal is 18\% protein. If the total interest on at 0.09 is 360 protein. How many pounds of each should be mixed together in order to get 360 lb that is \newline
        \textbf{Trans}: soybean meal is 18\% protein. cornmeal is 9\% protein. How many grams of each should be mixed together in order to get 360 grams of the mixture. \newline
        \textbf{DualCG}: A chemist has a solution of 360 solution and a 20\% solution of alcohol. How many liters of a solution that is the mixture that no solution to make a mixture that is 50\% alcohol.
		\\
		\midrule
		\midrule
		\textbf{Equ}: $ equ : x + y = 35 \  equ : x / y = 2 / 5$  \newline \textbf{MT}: total of 1600 people work for \textcolor{purple}{a company}. \textcolor{purple}{The ratio of male to female employees} is 3 : 5. How many more \textcolor{purple}{females than males} are there in the company. \newline
		\textbf{Ours}: The \textcolor{blue}{ratio of boys to girls} in at \textcolor{blue}{a certain school} is 5 : 2. If there are total 35 \textcolor{blue}{boys and girls}. how many of each are there.\newline
        \textbf{Seq2seq}: The school art club is having a exhibit. The ratio of the school paintings are in two parts is 2 / 5 of the number. What is the number ?\newline \textbf{SeqGAN}: A carpet is 3 times as many more than the other. The total value is 3.\newline
        \textbf{DeepGCN}: The ratio of the larger of the two numbers is 35. The ratio of the smaller number of goals and the other is 5 / 2. What are the two numbers.  \newline \textbf{Trans}: Pat . 35 students and 5 questions. If the total of the students are seniors and 2 take both the total. how many of each. \newline
        \textbf{DualCG}: The sum of two numbers is 35. The larger number is 2 less than the smaller number. Find the larger number. \\
        
        \midrule
        \midrule
        \textbf{Equ}: $equ : 1 / 6 * x + 7 = 2 / 3 * x$  \newline \textbf{MT}: \textcolor{purple}{A truck's gas tank} is 1 / 4 full. After 15 \textcolor{purple}{gallons of gas} are \textcolor{purple}{added}, the tank is 7 / 8 full . What is the \textcolor{purple}{gallon capacity} of the gas tank? \newline \textbf{Ours}: \textcolor{blue}{Deandre's gas tank} is 2 / 7 full. After he \textcolor{blue}{buys} 6 \textcolor{blue}{gallons of gas}. it is 2 / 3 full. How many gallons can \textcolor{blue}{Deandre's tank hold}. \newline
     \textbf{Seq2seq}: If a sum of 1 and 7 / 3 of the other digit. What is the value of x .\newline
    \textbf{SeqGAN}: $<$UNK$>$ can paint a house in 6 hours . If the same time it takes 7 / 3 hours . How many days will it take to go 2 miles apart.\newline
    \textbf{DeepGCN}: The sum of the first three numbers is 7. the sum of the first number and the number is 7. the result is the same as when the result is one. Find the number  \newline \textbf{Trans}: 1 / 6 of a number is 7 / 2 of the number. Find the number.  \newline\textbf{DualCG}: If 1 / 6 of a number is 2 / 6. Find the number. \\
    \midrule
    \midrule
        \textbf{Equ}: $equ : 250 + 400 = x \ equ : 1625/x = y$ \newline \textbf{MT}: \textcolor{purple}{2 vehicles traveling} different directions. same \textcolor{purple}{start point} and time. one vehicle is \textcolor{purple}{60 mph}. the other is \textcolor{purple}{55 mph}. In how many hours will they be \textcolor{purple}{500 miles apart}.
        \newline \textbf{Ours}: \textcolor{blue}{Two cars leave Denver traveling} in opposite directions. One has a \textcolor{blue}{speed of 250 mph} and the other airplane \textcolor{blue}{averages 400 mph}. How many hours will the trip be \textcolor{blue}{1625 miles apart}. \newline
     \textbf{Seq2seq}: A $<$UNK$>$ of deposit costs \$ 400. 000 a t the end of the year. the total interest is \$ 1625 . 00. What is the total cost of the total.\newline
    \textbf{SeqGAN}: quotient of a certain number is 400. If the number of students in the first 250 is 400. What is the number.\newline
    \textbf{DeepGCN}: The car ran a t an average speed of 400 km per hour faster than the other. If the speed of a 400 mi / h faster. What was the speed of the plane in miles per hour.  \newline \textbf{Trans}: planes went to school a t a speed for the trip takes 250 mph for 400 hours. How long will the plane travel in the trip.  \newline\textbf{DualCG}: Joe received 250 miles for 250 miles . and gas a trip of 250 miles per hour for \$ 400 to the week. He drove 400 miles per hour faster . What was the average speed for the trip. \\
    
		\bottomrule
	\end{tabularx}}
	\caption{Four examples of math word problems generated by different models. Transformer is abbreviate to Trans. Equ and MT represents the equation and the matched MWP instance, respectively. Quantity-related attributes and predicates in the instance that are picked up and rewritten in the generated MWP are colored for better readability.}
	\label{case}
\end{table*}



\end{document}